%% file: PaperForReview.tex
\documentclass[10pt,twocolumn,letterpaper]{article}

\usepackage[pagenumbers]{cvpr} 

\usepackage{graphicx}
\usepackage{amsmath}
\usepackage{amssymb}
\usepackage{booktabs}

\input{math_commands.tex}

\usepackage{url}
\usepackage{graphicx}
\usepackage{tabularx}
\usepackage{sidecap}
\usepackage{multirow}
\usepackage{pifont}
\usepackage{makecell}
\usepackage{listings}
\usepackage{algorithm}

\usepackage[pagebackref,breaklinks,colorlinks]{hyperref}

\usepackage[capitalize]{cleveref}
\crefname{section}{Sec.}{Secs.}
\Crefname{section}{Section}{Sections}
\Crefname{table}{Table}{Tables}
\crefname{table}{Tab.}{Tabs.}

\def\cvprPaperID{6024} 
\def\confName{CVPR}
\def\confYear{2022}

\usepackage{xcolor}

\definecolor{codegreen}{rgb}{0,0.6,0}
\definecolor{codegray}{rgb}{0.5,0.5,0.5}
\definecolor{codepurple}{rgb}{0.58,0,0.82}
\definecolor{backcolour}{rgb}{0.95,0.95,0.92}

\lstdefinestyle{mystyle}{
    backgroundcolor=\color{backcolour},   
    commentstyle=\color{codegreen},
    keywordstyle=\color{magenta},
    numberstyle=\tiny\color{codegray},
    stringstyle=\color{codepurple},
    basicstyle=\ttfamily\footnotesize,
    breakatwhitespace=false,         
    breaklines=true,                 
    captionpos=b,                    
    keepspaces=true,                 
    numbers=left,                    
    numbersep=5pt,                  
    showspaces=false,                
    showstringspaces=false,
    showtabs=false,                  
    tabsize=2
}

\begin{document}

\title{Scale-Equivalent Distillation for Semi-Supervised Object Detection}

\author{Qiushan Guo\textsuperscript{1}, Yao Mu\textsuperscript{1}, Jianyu Chen\textsuperscript{2}, Tianqi Wang \textsuperscript{1}, Yizhou Yu\textsuperscript{1}, Ping Luo\textsuperscript{1}\\
\textsuperscript{1}The University of Hong Kong
\textsuperscript{2}Tsinghua University \\
{\tt\small \{qsguo,ymu,tqwang,yzyu,pluo\}@cs.hku.hk} 
{\tt\small jianyuchen@tsinghua.edu.cn}
}
\maketitle

\begin{abstract}
Recent Semi-Supervised Object Detection (SS-OD) methods are mainly based on self-training, i.e., generating hard pseudo-labels by a teacher model on unlabeled data as supervisory signals. Although they achieved certain success, the limited labeled data in semi-supervised learning scales up the challenges of object detection. We analyze the challenges these methods meet with the empirical experiment results. We find that the massive False Negative samples and inferior localization precision lack consideration. Besides, the large variance of object sizes and class imbalance (i.e., the extreme ratio between background and object) hinder the performance of prior arts. Further, we overcome these challenges by introducing a novel approach, Scale-Equivalent Distillation (SED), which is a simple yet effective end-to-end knowledge distillation framework robust to large object size variance and class imbalance. SED has several appealing benefits compared to the previous works.
(1) SED imposes a consistency regularization to handle the large scale variance problem. 
(2) SED alleviates the noise problem from the False Negative samples and inferior localization precision.
(3) A re-weighting strategy can implicitly screen the potential foreground regions of the unlabeled data to reduce the effect of class imbalance. 
Extensive experiments show that SED consistently outperforms the recent state-of-the-art methods on different datasets with significant margins. For example, it surpasses the supervised counterpart by more than 10 mAP when using 5\% and 10\% labeled data on MS-COCO.
\end{abstract}

\section{Introduction}
\begin{figure}[t]
\begin{center}
\includegraphics[trim=0cm 0cm 0cm 0cm,width=0.48\textwidth]{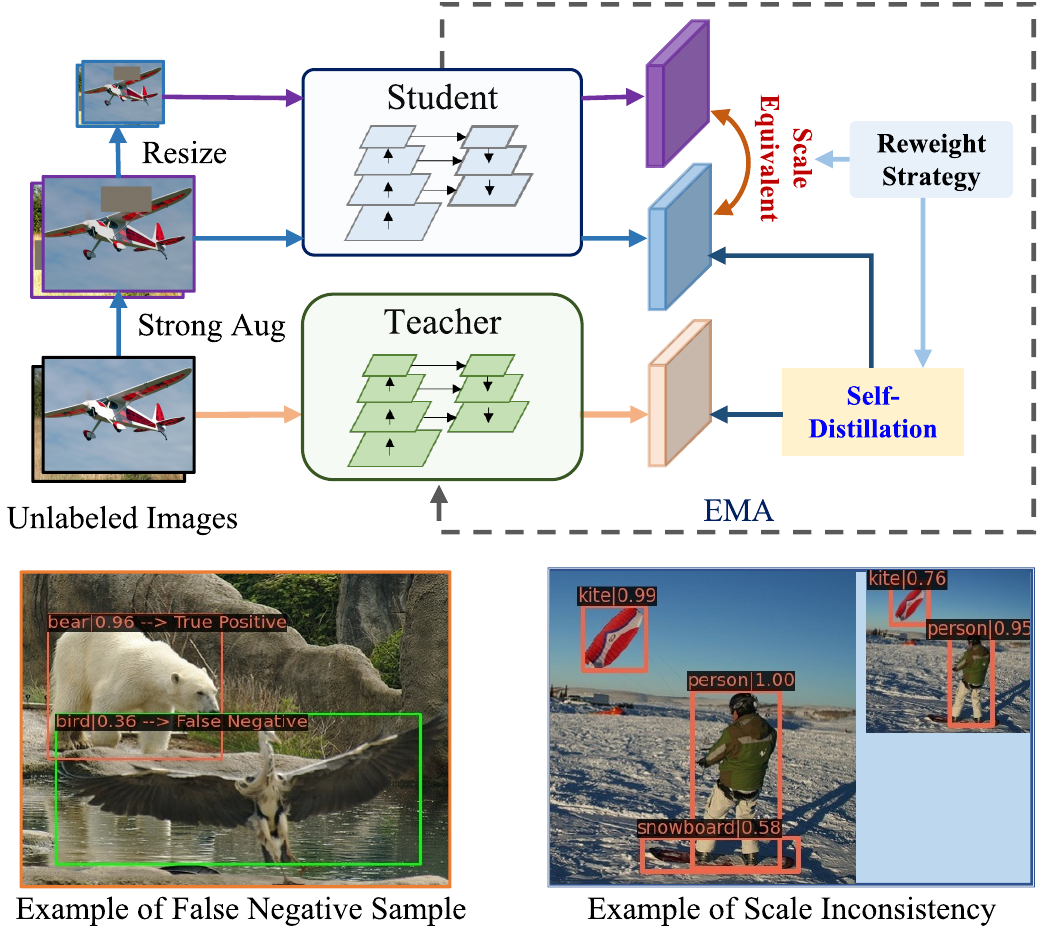}
\end{center}
\caption{The overall framework of SED. Our model improves the scale equivalence, which is critical for object detectors, by regularizing the consistency between different-sized images. Furthermore, the inherent False Negative sample noise is alleviated by self-distillation. A re-weighting strategy is adopted to solve the severe class imbalance problem. The bird in the left example is a False Negative sample when the threshold is set to 0.7. The right example shows the scale inconsistency of different-sized images.}
\label{intro}
\end{figure}
Deep neural networks achieve strong results under the supervised learning framework driven by large-scale datasets, such as ImageNet\cite{deng2009imagenet} (about 1.28 million labeled images). However, different from classification, object detection further involves locating objects with a bounding box. Therefore, the annotation for object detection is much more expensive, leading to labeled data remaining scarcely related to classification. 
Recently, Semi-Supervised Learning (SSL) for classification has received much attention \cite{tarvainen2017mean, berthelot2019mixmatch, xie2020self, sohn2020fixmatch}, whose results are comparable to the fully supervised model on ImageNet. However, Semi-supervised Object Detection (SS-OD) is more challenging than SSL on ImageNet classification. 
Recent SS-OD methods improve the performance by leveraging both the limited labeled data and the massive unlabeled data, but they suffer from the large variance of object sizes, massive False Negative samples and class imbalance problem, as illustrated in Fig.~\ref{intro}.

\begin{figure*}[t]
    \vspace{-20pt}
    \begin{center}
    \subfloat[][\label{obs_subfig_1}Instance Size Distribution]{%
      \includegraphics[trim=0.cm 0.cm 0.cm 0.cm,width=0.3\textwidth]{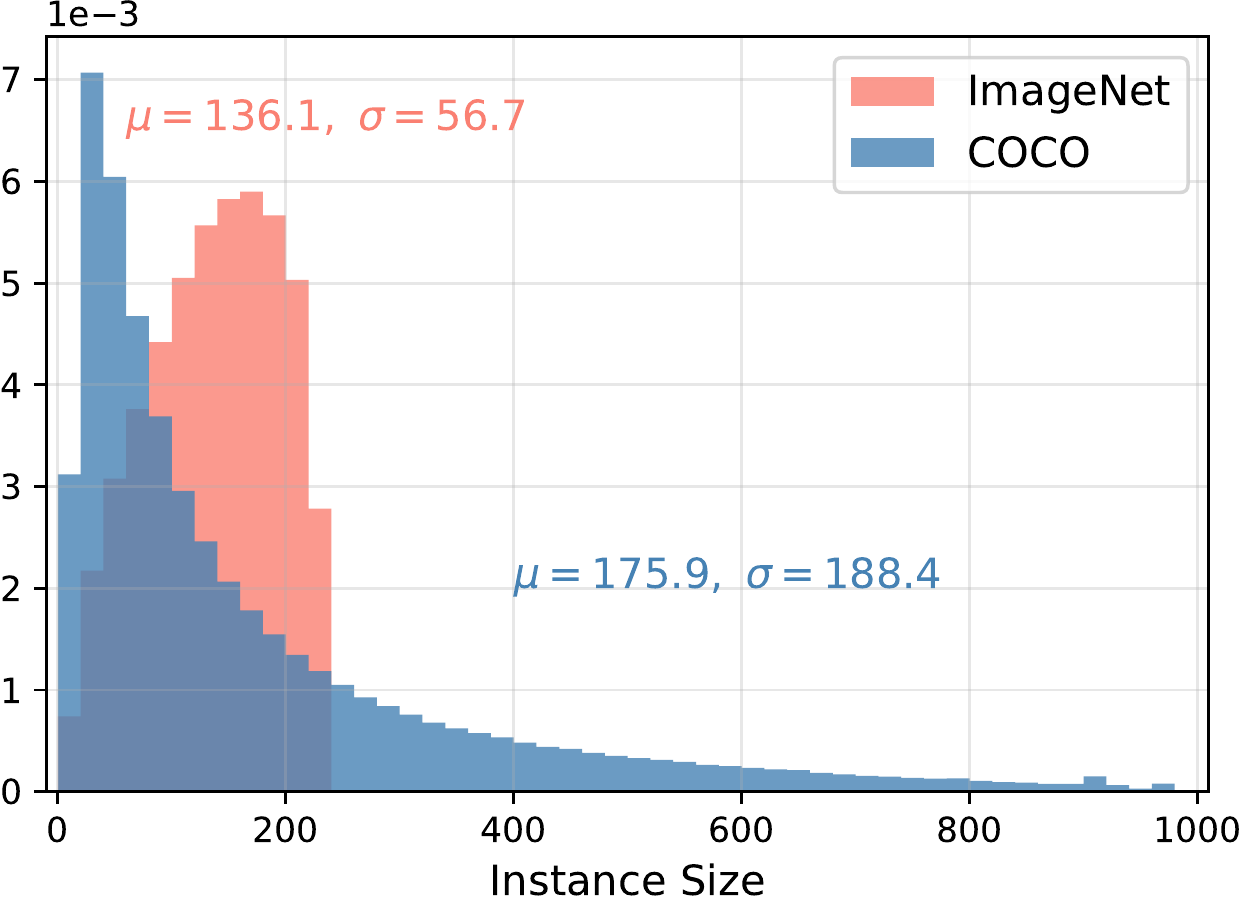}
     }
    \hfill
    \subfloat[][\label{obs_subfig_2}Score Distance Distribution]{%
      \includegraphics[trim=1.cm 0.3cm 0.cm 0.cm,width=0.32\textwidth]{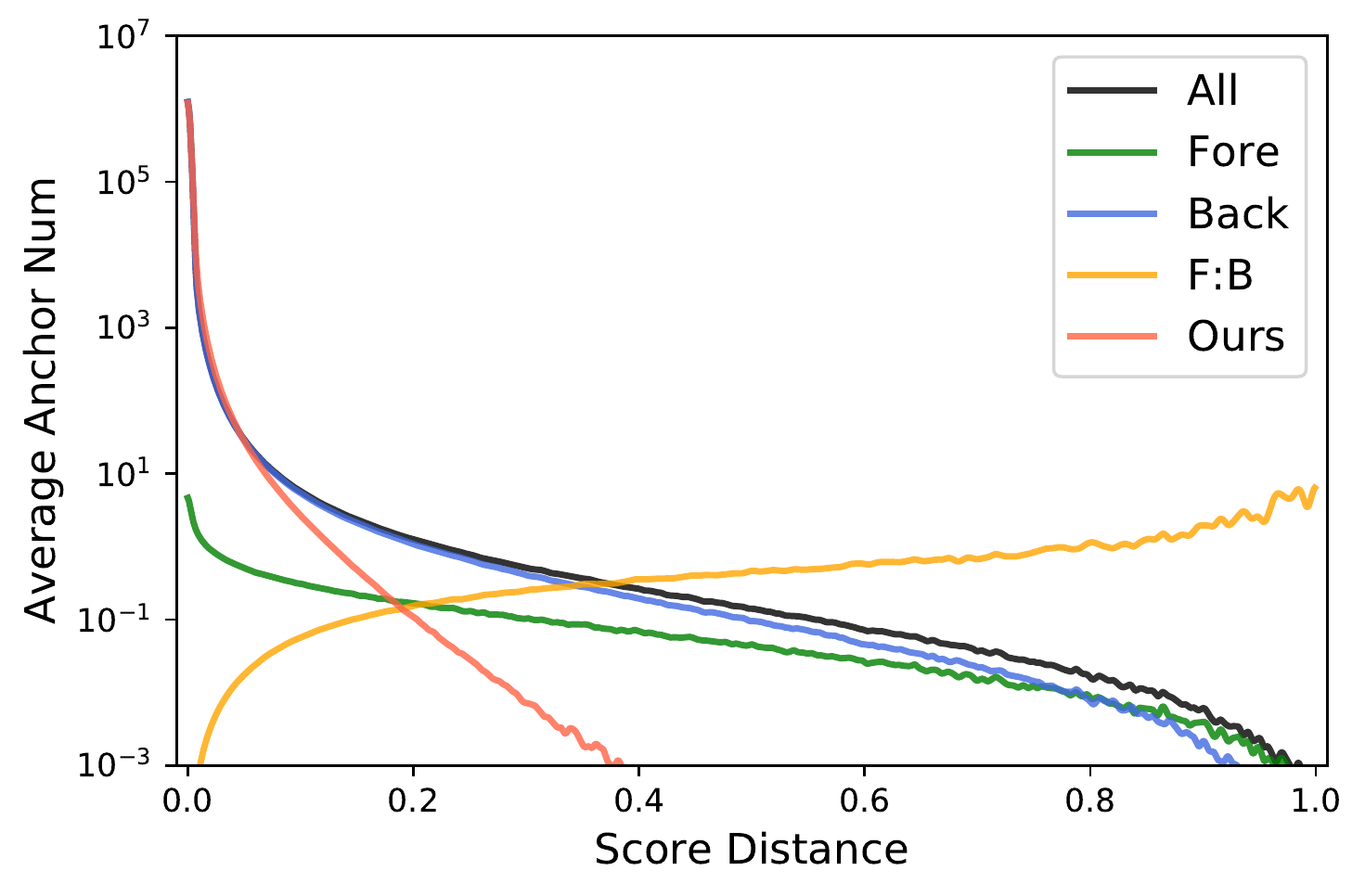}
     }
    \hfill
    \subfloat[][\label{obs_subfig_3}AP and AR {\it w.r.t} Score]{%
      \includegraphics[trim=0.7cm 0.cm 0.4cm 0.cm,width=0.3\textwidth]{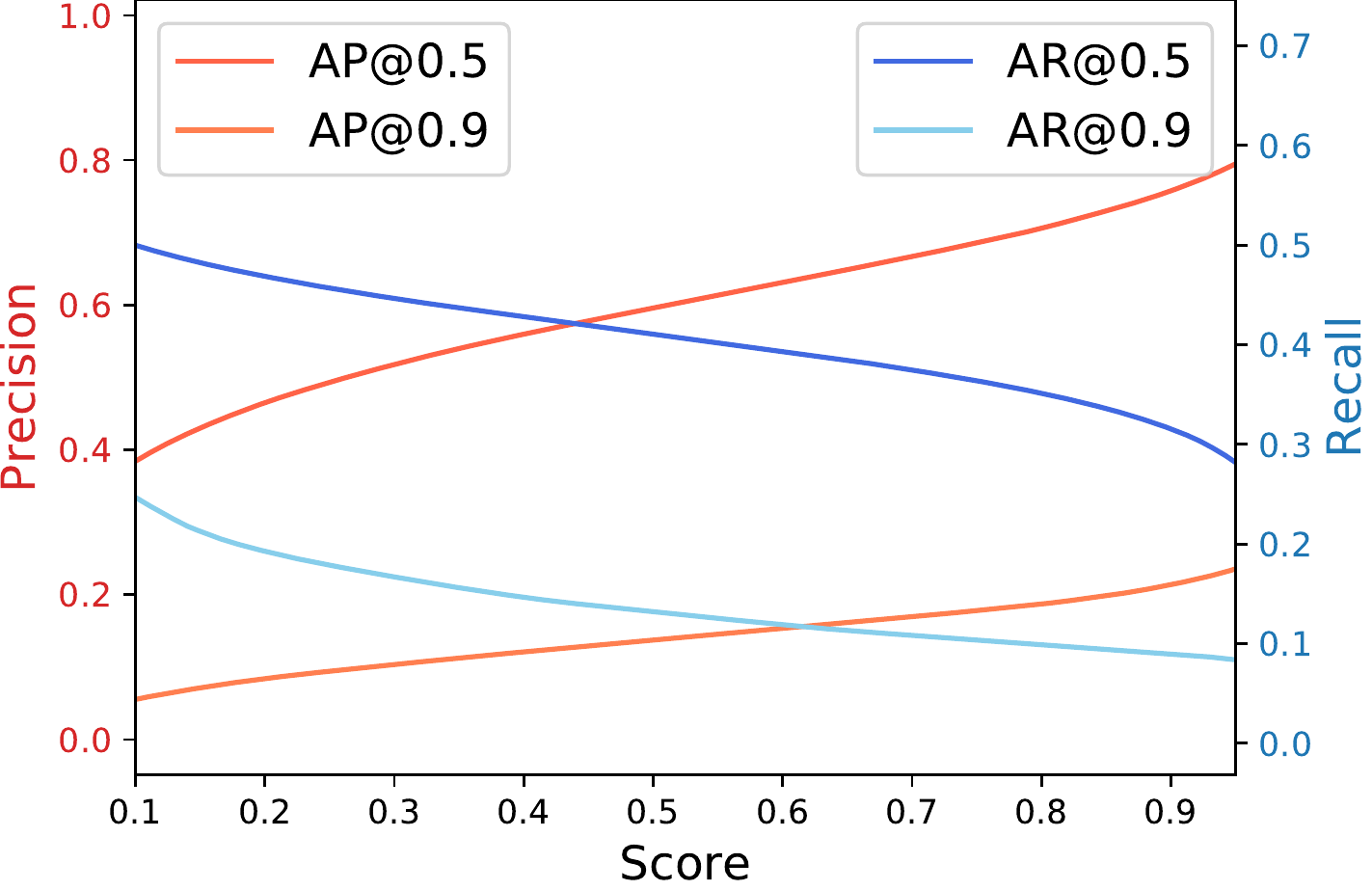}
     }
    \end{center}
    \vspace{-5pt}
    \caption{(a) For the COCO dataset, all the images are resized such that the short edge has 800 pixels while the long edge has less than 1333 pixels. For the ImageNet dataset, all the images are resized to 224$\times$224 to calculate the statistics. The scale of object is represented as the square root of the area. We discuss the typical training input size for ImageNet classification and COCO detection tasks. (b) All the scores are predicted on COCO {\it minival} dataset by the RetinaNet detector with FPN and ResNet 50 backbone, which is trained with 10\% COCO data. The score distance is the absolute difference between the predictions of the image in different sizes. The Y-axis is the average number of anchors per image. (c) We predict pseudo-label on the rest of COCO training data with a converged Faster-RCNN detector (with FPN and ResNet50 backbone), trained with 10\% COCO data. The low average recall and precision show that hard pseudo-label incur more noise with False Negative samples.}
    \label{Obs}
    \vspace{-5pt}
\end{figure*}

The scale of objects varies in a small range for the ImageNet classification model, whereas the scale variation of MS-COCO dataset \cite{lin2014microsoft} is large across object instances for the detector. As shown in Fig.~\ref{obs_subfig_1}, the standard deviation of the scale of instances in MS-COCO is 188.4 pixels, while that of ImageNet is 56.7 pixels (the square root of area). A detector is supposed to be scale consistent to object instances, which means that the predictions of an image in different sizes should be equivalent \cite{singh2018analysis, singh2018sniper}. However, the scale consistency has not been considered by the prior arts \cite{sohn2020simple, liu2021unbiased, zhou2021instant, yang2021interactive} in SS-OD.
We observe a discrepancy in the objectness score, as indicated in Fig.~\ref{obs_subfig_2}. The ratio of foreground anchor to background anchor increases as the score distance becomes large, which implies that the model detects an object instance while is blind to the instance in a different size.
This inconsistency is typically alleviated by the multi-scale inference ensemble, which increases the computational cost and requires complicated operations to fuse the results.

Besides, the performance of recent SS-OD methods \cite{sohn2020fixmatch, liu2021unbiased} is moderate in the high-data scenario as a consequence of the False Negative object instance and inferior localization precision. 
As illustrated in Fig.~\ref{obs_subfig_3}, the recall drops to 0.1 and 0.3 separately when IoU is set to 0.5 and 0.9, which indicates that most foreground instances are False Negative samples. The precision at IoU = 0.9 is less than 0.2, showing that the location of bounding boxes is not accurate enough. The False Negative object instances below the hard threshold cause a recognition inconsistency.

Another obstacle is that the foreground and background samples are highly imbalanced. The ratio of the foreground to background sample is about 1:25,000 for RetinaNet\cite{lin2017focal}. Due to the class imbalance problem, treating all regions equally \cite{tang2021humble} leads to the background samples contributing significantly to the gradient, as illustrated in Fig.~\ref{Grad}. Identifying foreground regions from the unlabeled data with the overwhelming background regions is challenging.

To overcome the challenges motioned above, we propose Scale-Equivalent Distillation (SED), a simple yet effective end-to-end semi-supervised learning framework for object detection. Since scale is an essential factor of the low-dimensional semantic manifolds, we design a scale consistency regularization across the prediction in different levels as a solution to the large object size variance. 
Moreover, as the noise from hard pseudo-label has detrimental effects on the recognition consistency, a self-distillation method is proposed to improve generalization performance without increasing the learnable parameters.
Due to the class imbalance problem, the overwhelming background samples diminish the effect of our method. We implement a re-weighting strategy to focus on the inconsistency among outputs in different levels and the discordance between the teacher and student detector. As a result, our re-weighting approach avoids selecting the potential foreground regions from the unlabeled data explicitly.

To evaluate the validation of SED, we conduct extensive experiments on benchmarks for object detection, Pascal VOC \cite{everingham2010pascal} and MS-COCO \cite{lin2014microsoft}. Our method surpasses the supervised counterpart by more than 10 mAP when using 5\% and 10\% labeled data on MS-COCO. Moreover, our method is tested with both one-stage and two-stage detector based on single feature map and feature pyramid. 

Our contributions are listed as follows:
(1) SED imposes a scale consistency regularization to overcome the large scale variance challenge. 
(2) SED alleviates the noise problem which arises from the False Negative samples and inaccurate bounding box regression.
(3) A re-weighting strategy can implicitly screen the potential foreground regions from unlabeled data to reduce the effect of class imbalance.

\section{Related Works}

{\bf Self-Training.}
Self-training methods first train a teacher model with the labeled dataset and then generate pseudo-labels for the unlabeled dataset. Finally, the student model is optimized with both the labeled data and pseudo-labeled data jointly. For the classification task, Self-training methods \cite{tarvainen2017mean, berthelot2019mixmatch, berthelot2020remixmatch, sohn2020fixmatch} perform well. However, Semi-Supervised Object Detection is more challenging than Semi-Supervised Image Classification on the balanced dataset. 
Some works \cite{liu2021unbiased, zhou2021instant} contribute to alleviating the noise problem brought by pseudo-label. Those methods attach additional modules on the two-stage detector to overcome the heavy overfitting problem on the foreground and background classification and refine the hard pseudo-label by ensemble methods. Nevertheless, methods based on hard pseudo-label have an inherent defect that False Negative object instances influence the consistency of recognition, especially whose scores are near the hard threshold. Humble Teacher \cite{tang2021humble} adopts soft pseudo-labels to avoid the recognition inconsistency but treat all the regions equally. Due to the extreme imbalance of foreground and background, the contribution of gradients from the two kinds of regions is quite different. UBT \cite{liu2021unbiased} adopts Focal Loss to alleviate the problem.
Unlike the existing works, our method generates soft pseudo-labels for unlabeled data in an online manner, and the re-weighting strategy automatically screens the potential foreground regions of the unlabeled data.

{\bf Consistency Regularization.}
Consistency-based Semi-supervised learning uses unlabeled data to stabilize the predictions under input or weight perturbations. 
For instance, two different views of the same image are supposed to have similar output.  
This class of methods \cite{samuli2017temporal, tarvainen2017mean, miyato2018virtual} does not generate pseudo-label but constrains the discrepancy between the outputs, which is known to help smooth the manifold \cite{oliver2018realistic}. For SS-OD, CSD \cite{jeong2019consistency} applies simple horizontal flip consistency regularization to train a detector to be robust to flip perturbations. The consistency loss fine-tunes the location of the predicted boxes but ignores the object scale perturbations, which are more common in datasets. In MS-COCO \cite{lin2014microsoft} detection dataset, the scale of the smallest and largest 10\% of object instances is 0.024 and 0.472, respectively.
Our method regularizes the predictions of different sizes to solve the large-scale variation. Furthermore, self-distillation \cite{furlanello2018born, zhang2018deep, guo2020online} benefits from the high-quality prediction of EMA teacher \cite{tarvainen2017mean}, and can be viewed as consistency regularization from the perspective of soft targets.

{\bf Pre-Training.}
In recent years, it has been a paradigm that pre-train backbone on a large-scale dataset, such as ImageNet \cite{deng2009imagenet} or JFT \cite{sun2017revisiting}, and fine-tune the model on the target dataset, which contains less training data. Large-scale dataset pre-training speeds up converge and helps improve generalization in the small data scenario\cite{he2019rethinking, zoph2020rethinking}, which is an extreme of semi-supervised learning. SimCLR \cite{pmlr-v119-chen20j} and MOCO \cite{he2020momentum} have been shown to build universal representation, which helps achieve a state-of-the-art result in the scenario of semi-supervised learning classification with 10\% ImageNet labeled data.
In this paper, we fine-tune with ImageNet pre-trained backbone as default for faster convergence and better results when we enter the low-data regime.

\vspace{-5pt}
\section{Scale-Equivalent Distillation} \label{ProblemDef}

\begin{figure*}[]
\begin{center}
\includegraphics[trim=0cm 0cm 0cm 0cm,width=0.85\textwidth]{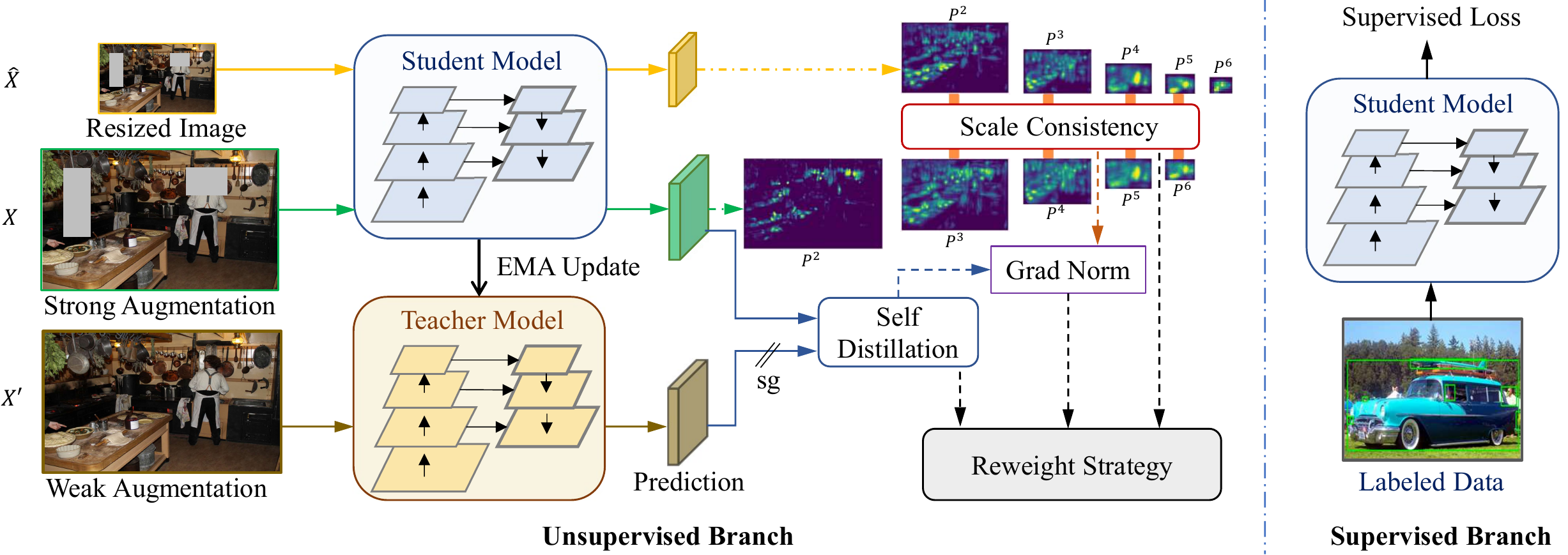}
\end{center}
\vspace{-5pt}
\caption{Details of our method. We take an example of a detector with FPN \cite{lin2017feature} to illustrate our method. $P^{2}$-$P^{6}$ are the prediction details.The supervised branch shares the Student Model with the unsupervised branch. \textit{sg} means the prediction of the Teacher Model is not optimized by the gradient. For Scale Consistency Regularization, the loss constrains the predictions from different levels.}
\label{overview}
\vspace{-5pt}
\end{figure*}
\textbf{Problem Definition.}
Semi-supervised learning is halfway between supervised and unsupervised learning.
More precisely, our model is trained with a labeled set $D_\text{s}=\{x_{i}^\text{s}, y_{i}^\text{s}\}_{i=1}^{N_\text{s}}$ and an unlabeled set $D_\text{u} = \{x_{i}^\text{u}\}_{i=1}^{N_\text{u}}$, where $x$ is image, $N_\text{s}$ and $N_\text{u}$ are the number of labeled and unlabeled images. For each supervised image $x_{i}^\text{s}$, the annotation $y_{i}^\text{s}$ is composed of both the location and category of the bounding boxes in image. 

\textbf{Overview.}
During training, Scale-Equivalent Distillation consists of two branches, the supervised and unsupervised branch, as illustrated in Fig.~\ref{overview}. The supervised branch is trained by following the normal procedure, like \cite{ren2015faster, lin2017focal}. The unsupervised branch is under a teacher-student framework, in which the teacher is implemented as an exponential moving average of the student. SED aims to predict consistently for the scale variants of input. In practice, the student processes the strongly augmented unlabeled images and resized images. The weakly augmented images are fed into the teacher network to predict soft pseudo-label. The scale consistency loss constrains the outputs of different-sized images. Meanwhile, the soft pseudo-label is set as the target of the strongly augmented images. The final loss is the weighted sum of the supervised and unsupervised loss, 
\begin{equation}
L = L_{\text{supervised}} + \frac{n_{\text{u}}}{n_\text{s}}(\lambda_\text{s} L_\text{scale} + \lambda_{d} L_\text{distill}),
\label{ALLLoss}
\end{equation}
where $n_\text{u}$, $n_\text{s}$ are the batch size of unlabeled data and labeled data. $L_\text{scale}$ and $L_\text{distill}$ are Scale Consistency Loss and Self-Distillation Loss. For two-stage detectors, the unsupervised losses are applied to both the RPN and RoI head.

\subsection{Scale Consistency Regularization}\label{ScaleSec}
Recognizing objects in different scales is a fundamental challenge in computer vision. Scale Consistency Regularization is proposed to optimize the detector to predict smoothly and consistently in scale dimension.
Typically, mainstream detectors under the feature pyramid network (FPN) framework outperform a single feature map counterpart, as the multi-scale feature representations are semantically strong. Therefore, we take an example for a single-stage detector with FPN to illustrate our method. Scale Consistency Regularization can be extended to the two-stage detectors and single feature map detectors. 

As indicated in Fig.~\ref{overview}, scale consistency loss regularizes predictions from images in different scales. To be more specific, the output class probability and bounding box regression of the $f$-th feature level, $r$-th row, $c$-th column and $d$-th anchor box are denoted as $P^{f,r,c,d}(X)$ and $R^{f,r,c,d}(X)$. Considering the memory and calculational cost, the resized image is downsampled to $\frac{1}{2^s}$ original size. Towards handling the large scale variation, the $s$ is uniformly selected from $\{1,2,...,S\}$, which also matches the sizes of feature maps in FPN and the label assignment rules. The resized image $\hat{X}$ and the original image $X$ are supposed to be predicted consistently for the corresponding levels. Precisely, the scale consistency loss is defined as 
\begin{equation}
\begin{split}
L_\text{scale}^{f} = &D_{\text{KL}}(\text{sg}(P^{f}(X)), P^{f'}(\hat{X}))\\
&+ D_{\text{KL}}(\text{sg}(P^{f'}(\hat{X})), P^{f}(X)) \\
&+ || R^{f}(X) - R^{f'}(\hat{X})||_{2},
\end{split}
\label{ScaleConsistencyLoss}
\end{equation}
where $f'$ equals $f-s$ and $\text{sg}$ is stop-gradient operator. For simplicity,  the $r,c,d$ coordinate is ignored in Eq.~\ref{ScaleConsistencyLoss}. For RPN and single-stage detector, all the anchor points are regularized for consistency; even some of them may not be assigned labels according to the simple IOU threshold matching strategy. In the second-stage detector framework, the proposals are first filtered by NMS and Top-K selection, which is also a default operation in the supervised branch \cite{ren2015faster, he2017mask} (typically 1000 proposals left for Faster-RCNN FPN). Then the coordinates of the proposals predicted on the resized image are scaled up by $2^s$ times to match the original image, and vice versa. The proposals from the image pair are simply concatenated as a new proposal set for refining bounding boxes and predicting classification scores. For the second stage of Faster-RCNN, all the predictions of the proposal pairs are regularized by scale consistency loss in a similar way as shown in Eq.~\ref{ScaleConsistencyLoss}. It is worth noting that, in implementing a two-stage detector, the RoI-Pooling operator may extract features from the same level for the proposal pair, which is slightly different from single-stage detectors. Nevertheless, this operation shares the same core idea that the detector is supposed to be scale consistent.

\subsection{Self-Distillation}
\begin{figure*}
\begin{center}
\includegraphics[trim=0.3cm 0cm 0.3cm 0cm,width=0.99\textwidth]{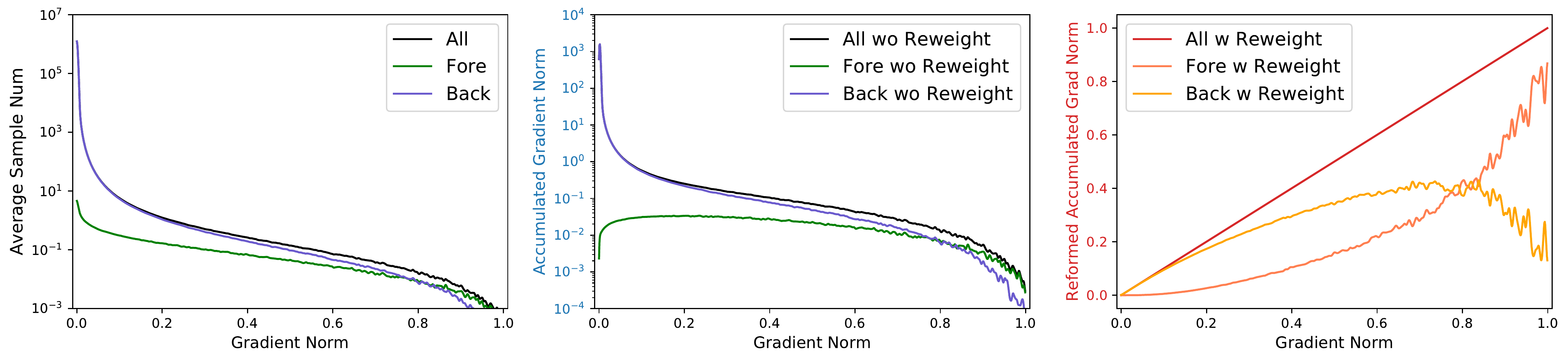}
\end{center}
\caption{The average sample is the average anchor number in a single image. The vanilla method is simply treating all samples equally. The samples with large gradients do not contribute significantly because the sample number is relatively small. Our re-weighting strategy focuses on the samples with large score discrepancies and linearizes the relationship between gradient contribution and score distance.}
\label{Grad}
\vspace{0pt}
\end{figure*}
Knowledge distillation improves generalization by replacing hard label supervision with soft label predicted by a stronger teacher model.
Based on the observation, the teacher model is implemented as an exponential moving average (EMA) of the detector, which is shown to produce a model with better generalization than the student model \cite{polyak1992acceleration,tarvainen2017mean}. The input of the teacher model is weakly augmented. Furthermore, the model is supposed to predict consistently for similar data points. The student model is input with the strongly augmented image to propagate label to neighbor points in the semantic manifold space. For simplicity, the strong augmentation is only composed of color transformation and Cutout \cite{devries2017improved}, which doesn't contain the geometric transformation. The self-distillation loss is formulated as 
\begin{equation}
\begin{split}
L^{i}_\text{distll} = D_{\text{KL}}(\text{sg}(P^{i}(X',\theta_t)), P^{i}(X,\theta_s))\\ 
+ || \text{sg}(R^{i}(X',\theta_t)) - R^{i}(X,\theta_s)||_{2},    
\end{split}
\label{WeightConsistencyLoss}
\end{equation}
where $i$ is the $i$-th anchor box, $X'$ is the weakly augmented image and $X$ is the strongly augmented image. $P$ and $R$ represent the classification score and bounding box regression same as in Eq.~\ref{ScaleConsistencyLoss}. The slowly progressing teacher model weights $\theta_t$ are updated from the student model weights $\theta_s$ every iteration,
\begin{equation}
\theta_t = \alpha \theta_t + (1-\alpha) \theta_s.
\label{EMATeacher}
\end{equation}
Self-Distillation loss constrains each anchor point for RPN and one-stage detector, similar to Scale Consistency Regularization. In the scenario of the two-stage detector, all the proposals are simply concatenated as a new proposal set. Similar to Scale Consistency Regularization, all the predictions of RoIs are regularized as Eq.~\ref{WeightConsistencyLoss}.

\subsection{Re-weighting Strategy}
One-stage object detection methods, like RetinaNet \cite{lin2017focal} and RPN \cite{ren2015faster}, face an extremely class imbalance during training. Due to the overwhelming background samples, most objectness scores are close to 0. Therefore, the KL divergence between the target and source distribution in Eq.~\ref{ScaleConsistencyLoss} and Eq.~\ref{WeightConsistencyLoss} is close to 0 for most anchor boxes. Simply averaging the unsupervised loss leads to the easy samples contributing significantly to the gradient, as illustrated in Fig.~\ref{Grad}. We aim to reduce the discrepancy between similar unlabeled inputs, especially for the potential foreground instances predicted with high objectness scores. In other words, the hard examples should contribute to the gradient more than the easy examples. Inspired by the Gradient Harmonizing Mechanism \cite{li2019gradient}, we re-weight the KL-divergence by the sample numbers in a gradient range to build a linear relationship between the gradient norm and the integral gradient contribution, as illustrated in Fig.~\ref{Grad}. Specifically, the gradient of the logits $z$ with KL divergence loss between probability vector $p$ and target probability vector $p'$ is $g = \sum_{i=1}^C |p_{i} - p'_{i}|$, 
where $C$ is the length of probability vector. Then a histogram is constructed by splitting the gradient range $[0,1]$ into $M$ bins equally. The number of samples in the $j$-th bin is denoted as $R_j$, and the index of the bin where gradient $g$ is located is defined as $idx(g)$. Finally, we have the loss function on $N$ samples:
\begin{equation}
L = \frac{1}{M}\sum_{i=1}^{N} \frac{D_{\text{KL}}(p'_i, p_i)}{R_{\text{idx}(g_i)}}.
\label{ReWeightLoss}
\end{equation}
As the main bottleneck is detecting objects from the background rather than regression, only the classification loss is re-weighted by the above strategy in scale consistency loss and self-distillation loss. Our goal is to enlarge the contribution from the samples with significant discrepancies. The other methods to solve the class imbalance problem may also improve the performance.

\begin{table}
\begin{center}
\vspace{5pt}
\begin{tabular}{lllll}
\toprule[1pt]
Method & Data & LR & Iter & AP$_{50}$  \\
\midrule
Supervised & VOC07 & 0.01 & 40k  & 74.30 \\
STAC \cite{sohn2020simple} & VOC07+12 & 0.001 & 180k & 77.45 \\
DGML \cite{wang2021data} & VOC07+12 & - & - & 78.60  \\
\midrule
UBT \cite{liu2021unbiased} & VOC07+12 & 0.01 & 180k & 77.37  \\
ISMT \cite{yang2021interactive} & VOC07+12 & - & - & 77.23  \\
IT \cite{zhou2021instant} & VOC07+12 & 0.01 & 180k & 78.30  \\
Ours & VOC07+12 & 0.01 & $\mathbf{40k}$ & $\mathbf{80.60}$ \\
\bottomrule[1pt]
\end{tabular}
\end{center}
\vspace{-5pt}
\caption{Results on Pascal VOC 2007 test set. For all the semi-supervised methods, Pascal VOC 2012 train set is treated as unlabeled data. Iter means the total training iterations. ``-''  means that the results or training details are missing in the source paper.}
\label{tab:VOC_Res}
\vspace{-5pt}
\end{table}

\begin{table*}
\vspace{-13pt}
\begin{center}
\begin{tabular}{lllllcl}
\toprule[1pt]
 Method & \multicolumn{3}{c}{Data Percent} & LR & Iteration & Stages   \\ 
\cline{2-4}  
 &  5\% & 10\% & 100\% & & & \\
\midrule
Supervised & 18.47 & 23.86 & 38.40 & 0.02 & 180k & - \\
STAC \cite{sohn2020simple} & 24.38{(+5.91)} & 28.64{(+4.78)} & - & 0.01 & 180k & Two \\
Unbiased Teacher \cite{liu2021unbiased}  & 27.84{(+9.37)} & 31.39{(+7.53)} & - & 0.01 & 180k & Single \\
Instant Teacher \cite{zhou2021instant} & 26.75{(+8.28)} & 30.40{(+6.54)} & 40.20{(+1.80)} & 0.01 & 180k & Single \\
Interactive Teacher \cite{yang2021interactive} & 26.37{(+7.90)} & 30.53{(+6.67)} & 39.64 {(+1.24)} & - & - & Single \\
Multi-Phase Learning \cite{wang2021data} & - & - & 40.30 {(+1.90)} & - & - &  Three \\
{\bf Ours} & $\mathbf{29.01}${\bf(+10.54)} & $\mathbf{34.02}${\bf(+10.16)} & $\mathbf{41.50}${\bf(+3.10)} & 0.01  & 180k & Single \\
\midrule
Supervised & - & - & 40.20 & 0.02 & 270k & - \\
STAC \cite{sohn2020simple} & - & - & 39.21{(-0.99)} & 0.01 & 540k & Two \\
Unbiased Teacher \cite{liu2021unbiased} & - & - & 41.30{(+1.10)} & 0.01 & 270k & Single \\
{\bf Ours} & - & - & $\mathbf{43.40}${\bf(+3.20)} & 0.02 & 270k & Single \\
\bottomrule[1pt]
\end{tabular}
\end{center}

\caption{Results on MS-COCO 2017 val set. For 5\% and 10\% protocols, the results are the mean over 5 data folds. Stages are the number of training phases. For example, STAC has two stages: train a teacher model first to hard pseudo-label and train a student model with both labeled and pseudo-labeled data. ``-'' means that the results or training details are missing in the source paper.}
\label{tab:COCO_Res}
\vspace{-5pt}
\end{table*}

\section{Experiments}
\textbf{Datasets.}
We mainly verify the validity of our method on the challenging objective detection dataset MS-COCO \cite{lin2014microsoft}, which contains 80 object categories with about 118k images for training and 5k images for validation. For a fair comparison, we follow the experimental setup as in the previous works \cite{sohn2020simple, liu2021unbiased, zhou2021instant, tang2021humble, wang2021data}. In particular, there are three experimental settings: 
(1) \textit{PASCAL VOC}: the VOC07 \cite{everingham2010pascal} \textit{trainval} set is used as the labeled dataset and the VOC12 \textit{trainval} set is used as the unlabeled dataset , as described in Sec.\ref{ProblemDef}. The performance is evaluated on the VOC07 test set. VOC07 \textit{trainval} and VOC12 \textit{trainval} contains 5,011 and 11,540 images respectively, resulting in a roughly 1:2 ratio of labeled data to unlabeled data.
(2) \textit{COCO-standard}: we randomly sample 5 and 10\% of MS-COCO 2017 training data as the labeled dataset and treat the rest of the training data as the unlabeled dataset. 
As the COCO train dataset contains 118k images and is class-imbalanced, some categories are composed of less than 500 instances. When data percent is 0.5\% and 1\%, there are only less than 5 instances in the labeled dataset for these categories. This setting is more like few-shot learning than semi-supervised learning. Therefore, we do not report the performance. 
For the 100\% data training setting, the whole training set is used as the labeled dataset, and the additional 123k unlabeled images are used as the unlabeled dataset. The model is tested on the MS-COCO 2017 validation set.
(3) \textit{COCO-35k}: we use the 35k subset of MS-COCO 2014 validation set as the labeled dataset and the 80k training set as the unlabeled dataset. The result is reported on the MS-COCO 2014 minival set.

\textbf{Implementation Details.}
Following STAC \cite{sohn2020simple}, we use Faster-RCNN \cite{ren2015faster} with FPN \cite{lin2017feature} and ResNet-50 backbone as our default object detector. The weights of the backbone are initialized by the corresponding ImageNet-Pretrained model, which is a default setting in the existing works \cite{sohn2020simple, jeong2019consistency, liu2021unbiased, zhou2021instant}. The stem and first stage of the backbone are frozen, and all BatchNorm layers are in {\it eval} mode. The weak data augmentation only contains random resize from (1333, 640) to (1333, 800) and random horizontal flip. The strong data augmentation comprises random Color Jittering, Grayscale, Gaussian Blur, and Cutout \cite{devries2017improved}, without any geometric augmentation. More training and data augmentation details are in the Appendix.

\subsection{Results}
\textbf{Pascal VOC}. 
In Tab.~\ref{tab:VOC_Res}, our method outperforms both previous multi-stage methods and single-stage methods by a large margin. Our model achieves 80.6\% AP with 6.3\% gain from additional VOC2012 data.
In the meantime, our proposed method requires fewer training iterations, showing that our approach is effective yet efficient. Besides, our augmentation only contains color transformation without any geometric transformation or strong regularization, such as Mixup \cite{zhang2017mixup} and DropBlock \cite{ghiasi2018dropblock}.

\textbf{COCO-standard}. 
Given the whole training set, our method even further improves the strong baseline by 3.2 mAP. 
For a fair comparison, the learning rate and training iterations are listed in Tab.~\ref{tab:COCO_Res}. Our method surpasses the previous methods under different settings of the ratio of labeled data to unlabeled data, from roughly 1:1 to 1:20, on the class-imbalanced MS-COCO dataset. Note that UBT uses Focal Loss to handle the class imbalance issue among ground truths, while we adopt the original Faster-RCNN implementation, standard cross-entropy loss. Our method focuses on the imbalance problem between foreground and background, which is more general in practice. Especially, SED achieves more than 10 mAP improvements against the supervised baseline when using 5\% and 10\% labeled MS-COCO data. With 10\% labeled data, the performance of SED is comparable to the fully supervised baseline model. 

\textbf{COCO-35k}.
MS-COCO 2014 minival set is identical to MS-COCO 2017 val set. Tab.~\ref{tab:35k_Res} shows that our method even outperforms the Oracle result with only 35k labeled data, benefiting from the scale consistency regularization, self-distillation, and strong augmentation. The promising result indicates that the semi-supervised method can achieve a comparable result to a fully supervised counterpart.

\begin{table}

\begin{center}
\resizebox{\linewidth}{!}{
\begin{tabular}{cccccc}
\toprule[1pt]
Method & SUP & DD\cite{radosavovic2018data} & DGML\cite{wang2021data} & Oracle  & Ours  \\
\midrule
mAP & 31.3 & 33.1 & 35.2 & 37.4 & $\mathbf{38.1}$ \\
\bottomrule[1pt]

\end{tabular}
}
\end{center}
\vspace{-8pt}
\caption{Results on MS-COCO 2014 minival set. SUP is to train the model only with the labeled data. Oracle means treating all the 115k images as labeled data and training with only the supervised loss.}
\vspace{-10pt}
\label{tab:35k_Res}
\end{table}

\subsection{Ablation Study}
{\bf Scale Consistency Regularization} constrains the discrepancy between the predictions of images of different sizes. By comparing the second row in Tab.~\ref{tab:Ab_Res} with baseline, we find that Scale Consistency Regularization improves about 3 mAP without our re-weighting strategy, naively averaging the loss across the anchor boxes and RoIs. Although suffering from the class imbalance problem, Scale Consistency Regularization is promising. Fig.~\ref{obs_subfig_2} shows that the discordance between different sizes is alleviated. 
\begin{table}

\begin{center}
\begin{tabular}{c|cccc|c}
\toprule[1pt]
 Method & SCR &  \multicolumn{2}{c}{Self-Distill} & Reweight & mAP   \\ 
 \cline{3-4}
  &  & Hard  & Soft  & & \\
\midrule
SUP &  &  &  &  & 23.86 \\
\midrule
\multirow{6}{*}{Ours} & \ding{52} &         &         &         & 26.80   \\
                      & \ding{52} &         &         & \ding{52} & 30.10   \\
                      &           &         & \ding{52} &         & 29.80   \\
                      &           &         & \ding{52} & \ding{52} & 31.40   \\
                      & \ding{52} & \ding{52} &         & \ding{52} & 29.50   \\
                      & \ding{52} &         & \ding{52} & \ding{52} & 34.00   \\
\bottomrule[1pt]
\end{tabular}
\end{center}
\vspace{-5pt}
\caption{The ablative results on MS-COCO 2017 val set. The models are trained with 10\% labeled and 90\% unlabeled MS-COCO train 2017 split. The SCR represents the scale consistency regularization. We test the self-distillation with two types of target: hard target and soft target.}
\label{tab:Ab_Res}
\vspace{-8pt}
\end{table}

{\bf Self-Distillation with Soft Target} surpasses the hard pseudo-label counterpart over 4.5 mAP, which demonstrates that the quality of hard pseudo-label is inferior. Self-Distillation gains about 6 mAP against the baseline individually. The soft target method benefits from fewer False Negative samples and the structural information via knowledge distillation. Furthermore, our approach based on soft target is threshold-free, which is simpler and easier to transfer to other datasets.

{\bf Re-weighting Strategy} focuses on the anchor or RoI pairs with large discrepancies and transforms the relationship between gradient contribution and score distance to linearity. The results of Scale Consistency Regularization and Self-Distillation with Soft Target are increased by 3.3 mAP and 1.6 mAP separately. For Faster-RCNN, our re-weighting strategy still takes effect even though the RoIs are predicted after NMS and Top-K selection operation, increasing the foreground to background sample ratio.

\subsection{Discussion}
{\bf How to Extend SED to Other Detectors.} Most detectors (e.g. RetinaNet, Faster-RCNN) assign foreground labels to the ``anchor box" according to a similar rule, the Intersection-over-Union (IoU) threshold criterion. For DETR \cite{carion2020end}, a single feature map detector, we match the predictions of input in different views according to Hungarian algorithm, where the pair-wise matching cost is defined as:
$L_\text{match} = D_\text{JS}(p_1, p_2)  +\lambda L_\text{IoU}(b_1,b_2)$,
where $D_\text{JS}(p_1, p_2)$ is JS-Divergence between the probability vectors and $L_\text{IoU}$ is GIoU loss \cite{rezatofighi2019generalized}.
According to the above analysis, our method can be extended to RetinaNet and DETR with different backbones. The results in Tab.~\ref{tab:SSD} demonstrate that our method is valid for different classes of the detector.

\begin{table}
\begin{center}
\begin{tabular}{lcc|c}
\toprule[1pt]
 Model & Retina w R50 & Retina w R18 & DETR w R50  \\ 
\midrule
SUP & 23.6 & 21.5 & 64.9 \\
Ours & 33.0 & 31.4 & 69.3  \\
\bottomrule[1pt]
\end{tabular}
\end{center}
\vspace{-5pt}
\caption{For RetinaNet, the experiments are conducted on MS-COCO set with 10\% labeled training data. Due to the extremely long training epoch of DETR, we report the result on Pascal VOC 2007 test set. Both supervised and our DETR are trained for 300 epochs.}
\label{tab:SSD}
\end{table}

\begin{table}
\begin{center}
\begin{tabular}{l|cc|c}
\toprule[1pt]
 Range & [640, 800] & [300, 1200] &  Ours [640, 800] \\ 
\midrule
Result & 31.4 & 32.0 & 34.0 \\
\bottomrule[1pt]
\end{tabular}
\end{center}
\vspace{-5pt}
\caption{The scale jittering results on MS-COCO 2017 val set. The models are trained with 10\% labeled and 90\% unlabeled MS-COCO train 2017 split. Range is the range of the short edge. The results show that the scale consistency loss is beyond large scale jittering augmentation.}
\label{tab:LSJ}
\vspace{-5pt}
\end{table}

{\bf Relationship with Large Scale Jittering.} The proposed scale consistency regularization is more than large scale jittering augmentation. The object of our method is $L = L(x) + L(x') +L_\text{scr}(x, x')$, while the object of a large scale jittering augmentation is $L =  L(x) + L(x')$, where $x$ and $x'$ are the input image in different views. The $L_\text{scr}$ is the scale consistency loss. The constraint of our method is stronger than large scale jittering augmentation. Thus we believe that the parameter space of local minimum is a subset of that of large scale jittering. Tab.~\ref{tab:LSJ} also shows that our method encourages the model to converge with less generalization error.

{\bf Relationship with Multi-Scale Testing.} 
Tab.~\ref{tab:MSTEST_Res} shows that the baseline models benefit from multi-scale testing by an ensemble with NMS (Threshold=0.5). The model trained with 10\% labeled data is increased by 2.0 mAP, and the fully supervised model (SUP 100\%) gets 1.5 mAP improvement. However, this improvement comes from the discrepancy between the predictions of images in different sizes. Moreover, the ensemble method also consumes $2.5\times$ more inference time than the single-scale testing method. Our method benefits less from multi-scale testing as a consequence of the proposed scale consistency regularization.
Our method significantly improves the single-scale testing performance, which has more practical value. 
\begin{table}[t]
\begin{center}
\begin{tabular}{lcccc}
\toprule[1pt]
 Model & \multicolumn{3}{c}{Image Size} & Ensemble \\ 
\cline{2-4}  
 &  480 & 800 & 1200 & \\
\midrule
SUP 10\% & 22.9 & 24.1 & 22.5 & 26.1{\tiny(+2.0)} \\
SUP 100\% & 33.7 & 37.4 & 36.8 & 38.9{\tiny(+1.5)}  \\
Ours 10\% & 31.5 & 34.2 & 33.0 & 34.8{\tiny(+0.6)}  \\

\bottomrule[1pt]
\end{tabular}
\end{center}
\vspace{-5pt}
\caption{Multi-Scale Testing on MS-COCO 2017 val set. The small gain indicates that the detector consistently predicts images in different sizes, which means robust to scale variance.}
\label{tab:MSTEST_Res}
\end{table}

\begin{figure}[]
\begin{center}
\includegraphics[trim=1cm 0cm 1cm 0cm,width=0.32\textwidth]{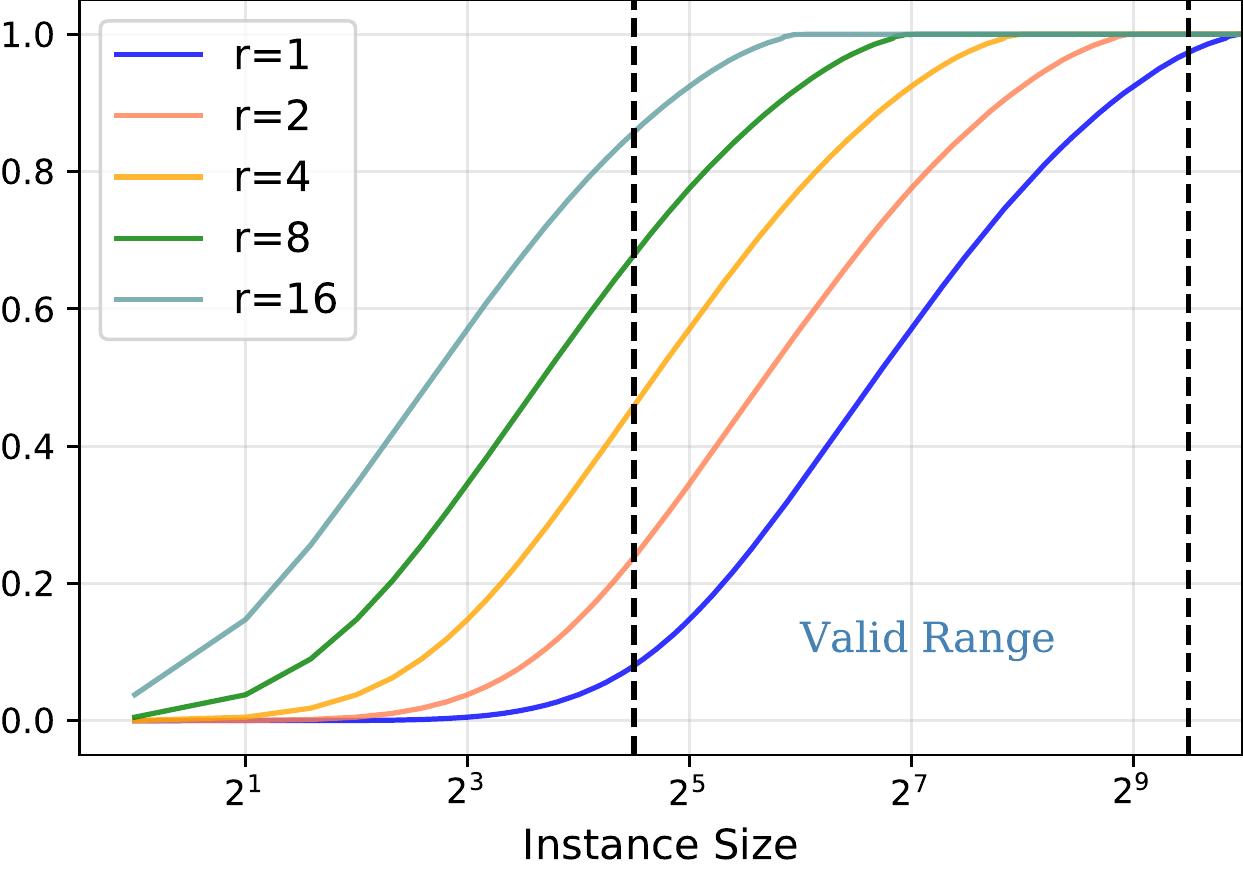}
\end{center}
\vspace{-10pt}
\caption{The CDF of instance size on MS-COCO train dataset.}
\label{fig:ValidRange}
\vspace{-10pt}
\end{figure}

{\bf Downsampling Rate in Scale Consistency Regularization.} 
As shown in Tab.~\ref{tab:DS_res}, the model achieves the best result when the downsampling rate is set to 2 (i.e. the $S$ in Sec.~\ref{ScaleSec} is set to 1.). The performance is inferior as the downsampling rate scales up, which means that regularizing the scale consistency with too small images is less effective. The anchor-based detector refines the prior boxes, which constrains the valid detection scale range (from 22.6 to 724.1 pixel$^2$, theoretically). Fig.~\ref{fig:ValidRange} shows that the fraction of instances in the valid range is highest when the downsampling rate is set to 2. All the models are trained with 10\% COCO training data, using RetinaNet with FPN and ResNet-18 backbone in Tab.~\ref{tab:DS_res}.

{\bf Exponential Moving Average (EMA) Rate of Teacher model.} In Eq.~\ref{EMATeacher}, the weight of the teacher is updated in an exponential moving average manner. The EMA update can be viewed as the average weight of the models in the past $\frac{\alpha}{1-\alpha}$ steps approximately. As the learning rate policy is {\it step}, which decays the learning rate by 0.1 at each milestone iteration, the performance of the teacher is inferior to the student model after switching the learning rate, which leads to the degradation of the student model. We observe the same appearance in UBT \cite{liu2021unbiased}, which sets the $\alpha$ to 0.9996 and adopts {\it step} learning rate policy. To alleviate the degradation, we propose to decay the EMA update rate at the same milestone iteration as the learning rate. The results in Tab.~\ref{tab:EMA} show that our {\it step} decay method and {\it cosine} decay method both surpass the baseline model. 

\begin{table}[t]
  \begin{minipage}{0.13\textwidth}
    \centering
    \resizebox{\textwidth}{!}{%
    \begin{tabular}{cc}
            \toprule[1pt]
            Rate & mAP \\
            \midrule
            1 & 23.0 \\
            2 & $\mathbf{26.1}$ \\
            4 & 25.2 \\
            8 & 23.1 \\
            16 & 21.1 \\
            \bottomrule[1pt]
        \end{tabular}
  
    }
  \caption{Results on COCO val set. Rate is the downsampling rate.}
  \label{tab:DS_res}
  \end{minipage}\hfill
  \begin{minipage}{0.28\textwidth}
      \centering
    \resizebox{\textwidth}{!}{%
    \begin{tabular}{cccc}
            \toprule[1pt]
            Start & End & Policy & mAP \\
            \midrule
            0.996 & 0.9 & Cosine & 33.0 \\
            0.99 & 0.9 & Step & $\mathbf{34.1}$ \\
            0.95 & 0.95 & None & 32.0 \\
            \bottomrule[1pt]
        \end{tabular}
    }
    \caption{Results on COCO val set. Start and End mean the initial EMA update rate and the target rate. Cosine policy is cosine annealing schedule. Our Step policy only decays once at the first milestone iteration.}
  \label{tab:EMA}
  \end{minipage}
\vspace{-5pt}
\end{table}

{\bf Compare Re-weighting strategy with other methods.} We conduct experiments by replacing our re-weighting strategy with OHEM (Online Hard Example Mining) and Focal Loss \cite{lin2017focal}. The vanilla method is training without any class balancing technique. The results in Tab.~\ref{tab:Focal} show that our method is effective.
\begin{table}
\begin{center}
\begin{tabular}{l|cccc}
\toprule[1pt]
 Method & Vanilla  & OHEM \cite{tang2021humble} & Focal \cite{lin2017focal} & Ours \\ 
\midrule
Result & 30.1 & 31.4 & 31.2 & 34.0 \\

\bottomrule[1pt]
\end{tabular}
\end{center}
\vspace{-12pt}
\caption{The comparison results of re-weight strategy on MS-COCO 2017 val set. The Faster-RCNN models are trained with 10\% labeled and 90\% unlabeled MS-COCO train 2017 split.}
\label{tab:Focal}
\vspace{-10pt}
\end{table}

\section{Conclusion}
In this work, we introduce a novel semi-supervised object detection framework based on the consistency regularization method. Our scale consistency regularization overcomes the large scale variance challenge and significantly improves the performance on single-scale testing. Further, SED alleviates the negative effect of False Negative samples and benefits from the structural information via knowledge distillation. The re-weighting strategy focuses on the potential fore-ground regions of the unlabeled data and linearizes the relationship gradient contribution and score distance. Experiments on MS-COCO and Pascal VOC show that Scale-Equivalent Distillation significantly improves the performance with different ratios of labeled data to unlabeled data and can be extended to different detector classes. 
Our framework is a holistic approach compatible with other semi-supervised methods, such as Mixmatch and Noisy student self-distillation. In addition, Scale-Equivalent Distillation framework could be further extended to other dense prediction tasks, like instance segmentation, joint human parsing, and post estimation. Our method has great potential to promote the development of semi-supervised learning and further reduce the dependence of labeled data with no negative social impact.


\begin{table*}[ht]
\begin{center}
\resizebox{\textwidth}{!}{
\begin{tabular}{cccc}
\toprule[1pt]
\multicolumn{4}{c}{Strong Augmentation} \\ 
\hline
Process         & Probability       & Parameters        & Details \\
\midrule
Color Jittering & 0.8               & brightness, contrast, saturation = 0.4, 0.4, 0.4 & 
\makecell[c]{
Brightness factor is chosen uniformly from [0.6, 1.4], \\
Contrast factor is chosen uniformly from [0.6, 1.4], \\
Saturation factor is chosen uniformly from [0.6, 1.4]
} \\
\hline
Grayscale & 0.2 & None & None \\
\hline
GaussianBlur & 0.5 & $\sigma \sim U(0.1, 2.0)$ & Gaussian filter kernel size is 23 \\
\hline
Cutout 1 & 0.7 & scale=(0.05, 0.2), ratio=(0.3, 3.3) & Randomly selects a rectangle region in an image \\
\hline
Cutout 2 & 0.5 & scale=(0.02, 0.2), ratio=(0.1, 6) & Randomly selects a rectangle region in an image \\
\hline
Cutout 3 & 0.3 & scale=(0.02, 0.2), ratio=(0.05, 8) & Randomly selects a rectangle region in an image \\
\bottomrule[1pt]
\end{tabular}
}
\end{center}
\vspace{-10pt}
\caption{Details of data augmentations.}
\label{tab:Aug}

\end{table*}

\appendix
\section{Appendix}

\subsection{Implementation of SED on DETR.}
Our method can be extended to DETR, a single feature map detector based on anchor-free label assignment rule. We match the predictions of input in different views according to Hungarian algorithm, where the pair-wise matching cost is defined as:
$L_\text{match} = D_\text{JS}(p_1, p_2)  +\lambda L_\text{IoU}(b_1,b_2)$,
where $D_\text{JS}(p_1, p_2)$ is JS-Divergence between the probability vectors and $L_\text{IoU}$ is GIoU loss \cite{rezatofighi2019generalized}. 
The python-style pseudo-code of matching algorithm is provided in Alg.~\ref{match}. The DETR model is trained with AdamW setting the transformer's learning rate to $10^{-4}$, the backbone's learning rate to $10^{-5}$, and weight decay to $10^{-4}$. The model is trained with a long schedule for 300 epochs and the learning rate is multiplied by 0.1 at 200 epochs. The other settings are the same as DETR \cite{carion2020end}. 

\subsection{Stronger augmentations.}
Geometric augmentations are common image data augmentations. Therefore, we further conduct experiments with stronger augmentations: color + geometric augmentations, to demonstrate the extendability of SED. We simply adopt the same geometric transformations in RandAug \cite{cubuk2020randaugment}, including {\it RandRotate}, {\it RandTranslation} and {\it RandShear}. We set the rand level to 5 and select only 1 transformation to apply. 
The results in Tab.~\ref{tab:VOC_Res} show that additional geometric augmentations lead to incremental improvement.

\begin{algorithm}[t]
\caption{Matching Pseudocode, PyTorch-like}
\label{alg:code}
\definecolor{codeblue}{rgb}{0.25,0.5,0.5}
\definecolor{codekw}{rgb}{0.85, 0.18, 0.50}
\lstset{
  backgroundcolor=\color{white},
  basicstyle=\fontsize{7.5pt}{7.5pt}\ttfamily\selectfont,
  columns=fullflexible,
  breaklines=true,
  captionpos=b,
  commentstyle=\fontsize{7.5pt}{7.5pt}\color{codeblue},
  keywordstyle=\fontsize{7.5pt}{7.5pt}\color{codekw},
}
\begin{lstlisting}[language=python]
def hungarian_match(cls_score_1, cls_score_2, bbox_pred_1, bbox_pred_2, cls_weight, iou_weight):
    # cls_score: [bs, num_query, c]
    # bbox_pred: [bs, num_query, 4]
    cls_dist = JSCost(cls_score_1, cls_score_2)
    iou_dist = IoUCost(bbox_pred_1, bbox_pred_2)
    cost= cls_dist*cls_weight + iou_dist*iou_weight
        
    bs = cost.shape[0]
    col_inds = []
    for i in range(bs):
        col_ind = linear_sum_assignment(cost[i])
        col_inds.append(col_ind)
    return col_inds
\end{lstlisting}
\label{match}
\end{algorithm}

\begin{table}[t]
\begin{center}
\vspace{-5pt}
\begin{tabular}{llll}
\toprule[1pt]
Method & Data &  AP & Augmentation  \\

\hline
Supervised & VOC07 &  74.3 & - \\
STAC \cite{sohn2020simple} & VOC07+12 &  77.45 & C, G\\
DGML \cite{wang2021data} & VOC07+12 & 78.60 & - \\
\hline
UBT \cite{liu2021unbiased} & VOC07+12 &  77.37 & C \\
ISMT \cite{yang2021interactive} & VOC07+12 & 77.23 & C, DropBlock \\
IT \cite{zhou2021instant} & VOC07+12 & 78.30 & C, Mixup, Mosaic \\
Ours & VOC07+12 &  $\mathbf{80.60}$ & $\mathbf{C}$ \\
Ours & VOC07+12 &  $\mathbf{81.44}$ & $\mathbf{C, G}$ \\
\bottomrule[1pt]
\end{tabular}
\end{center}
\vspace{-15pt}
\caption{Results on Pascal VOC 2007 test set.  $AP_{50}$ is reported. ``-''  means that the training details are missing in the source paper.}
\label{tab:VOC_Res}
\vspace{-10pt}
\end{table}

\subsection{Implementation and Training Details.}

Our implementation is based on MMDetection framework \cite{chen2019mmdetection}.  The default detector is set as Faster-RCNN \cite{ren2015faster} with FPN \cite{lin2017feature} and ResNet-50 \cite{he2016deep} for a fair comparison with prior works \cite{sohn2020simple, yang2021interactive, zhou2021instant, liu2021unbiased, wang2021data}. {\bf Code will be released.}

{\bf Training Details.} 
The weights of the backbone are first initialized by the corresponding ImageNet-Pretrained model, which is a default setting in existing works \cite{sohn2020simple, jeong2019consistency, liu2021unbiased, zhou2021instant}. All the models are trained with learning rate starting at 0.01. The learning rate drops by 0.1 at the 120k and 160k iteration for 180k training schedule as default. We set the weight decay to 0.0001, batch size to 16, and the momentum is 0.9 for SGD optimizer. Like \cite{liu2021unbiased}, we separate 5k/10k/12k/90k iterations from the whole process as the burn-in phase for 5\%/10\%/35k/100\% data protocols. For verifying the effectiveness of our method, we simply set the $\lambda_{s}$ and $\lambda_{d}$ in Eq.~1 as 0.5 and 1 separately. The EMA update rate starts with 0.99 and steps to 0.9 at the 120k iteration, aligned with the learning rate decay policy.

{\bf Data Augmentation.}
As shown in Tab.~\ref{tab:Aug}, the weak data augmentation only contains random resize from (1333, 640) to (1333, 800) and random horizontal flip with a probability of 0.5. The strong data augmentation is composed of random Color Jittering, Grayscale, Gaussian Blur, and Cutout \cite{devries2017improved}, without any geometric augmentation.


\input{PaperForReview.bbl}
{\small
\bibliographystyle{ieee_fullname}
\bibliography{PaperForReview}
}

\clearpage

\end{document}

%% file: math_commands.tex
\usepackage{amsmath,amsfonts,bm}

\def\eqref#1{equation~\ref{#1}}

\def\1{\bm{1}}

\DeclareMathAlphabet{\mathsfit}{\encodingdefault}{\sfdefault}{m}{sl}
\SetMathAlphabet{\mathsfit}{bold}{\encodingdefault}{\sfdefault}{bx}{n}

